\def\year{2021}\relax
\documentclass[letterpaper]{article} %
\usepackage{aaai21}  %
\usepackage{times}  %
\usepackage{helvet} %
\usepackage{courier}  %
\usepackage[hyphens]{url}  %
\usepackage{graphicx} %
\usepackage{natbib}  %
\usepackage{caption} %
\usepackage[utf8]{inputenc} %
\usepackage{url}            %
\usepackage{booktabs}       %
\usepackage{amsfonts}       %
\usepackage{nicefrac}       %
\usepackage{microtype}      %
\usepackage{epsfig}
\usepackage{graphicx}
\usepackage{amsmath}
\usepackage{amssymb}
\usepackage{gen symb}
\usepackage[dvipsnames]{xcolor}
\usepackage{colortbl}
\usepackage{subcaption}
\usepackage{verbatim}
\usepackage{mathtools}
\usepackage{multirow}
\usepackage{floatrow}
\usepackage{etoolbox}

\providetoggle{showcomments}									%
\settoggle{showcomments}{false} 								%

\iftoggle{showcomments}{%
\newcommand{\todo}[1]{\textcolor{blue}{\textbf{TODO:} #1}}

\newcommand{\michael}[1]{\textcolor{orange}{\textbf{Michael:} #1}}
\newcommand{\new}[1]{\textcolor{Fuchsia}{#1}}
\newcommand{\vitto}[1]{\textcolor{red}{\textbf{Vitto:} #1}}
\newcommand{\jasper}[1]{\textcolor{magenta}{\textbf{Jasper:} #1}}
\newcommand{\att}[1]{\textcolor{red}{#1}}
\newcommand{\scratch}[1]{\textcolor{red}{\sout{#1}}}
}{%
\newcommand{\todo}[1]{}

\newcommand{\michael}[1]{}
\newcommand{\vitto}[1]{}
\newcommand{\jasper}[1]{}
\newcommand{\new}[1]{#1}
\newcommand{\att}[1]{\textcolor{black}{#1}}
\newcommand{\scratch}[1]{}
}
\newfloatcommand{capbtabbox}{table}[][\FBwidth]  %
\renewcommand{\vec}[1]{\boldsymbol{\mathbf{#1}}}

\newcommand{\auxiliarytask}{auxiliary task}
\newcommand{\auxiliarytaskhead}{auxiliary task head}

\newcommand{\rp}{rotation prediction}
\newcommand{\dcc}{discriminating common classes}
\newcommand{\iiw}{identical initial weights}
\newcommand{\RP}{RP}
\newcommand{\DCC}{DCC}
\newcommand{\IIW}{IIW}

\newcommand{\targettask}{target task}
\newcommand{\taskhead}{task head}

\newcommand{\targettaskhead}{target task head}
\newcommand{\classificationhead}{classification head}
\newcommand{\featureextractor}{feature extractor}
\newcommand{\jointtraining}{joint training}
\newcommand{\incrementaltraining}{incremental training}
\newcommand{\para}[1]{\par\vspace{2pt}\noindent\textbf{#1}}

\newcommand{\uc}{\expandafter\MakeUppercase}
\newcommand{\accuracy}{recombination accuracy}
\newcolumntype{H}{>{\setbox0=\hbox\bgroup}c<{\egroup}@{}}
\def\dataset{\mathcal{D}}

\def\bTheta{\vec{\Theta}}
\def\bPhi{\vec{\Phi}}

\def\bx{\vec{x}}
\def\by{\vec{y}}

\def\bs{\vec{s}}

\def\indicator{\vec{1}}

\newcommand{\ie}{\textit{i.e.} }
\newcommand{\eg}{\textit{e.g.} }
\newcommand{\vs}{\textit{vs.} }

\newcommand{\wrt}{\textit{w.r.t.}}

\title{Towards Reusable Network Components by Learning Compatible Representations}

\iftrue
\author {
    Michael Gygli, %
    Jasper Uijlings, %
    Vittorio Ferrari %
    \\
}
\affiliations {
    Google Research \\
   gyglim@google.com, jrru@google.com, vittoferrari@google.com
}
\else
\author{
PaperID: 9942
\vspace{-25pt}
}
\fi
\begin{document}
\maketitle

\begin{abstract}
This paper proposes to make a first step towards compatible and hence reusable network components. Rather than training networks for different tasks independently, we adapt the training process to produce network components that are compatible across tasks. In particular, we split a network into two components, a features extractor and a target task head, and propose various approaches to accomplish compatibility between them. We systematically analyse these approaches on the task of image classification on standard datasets.
We demonstrate that we can produce components which are directly compatible without any fine-tuning or compromising accuracy on the original tasks. Afterwards, we demonstrate the use of compatible components on three applications: Unsupervised domain adaptation, transferring classifiers across feature extractors with different architectures, and increasing the computational efficiency of transfer learning.
\end{abstract}

\begin{figure*}
\centering
\includegraphics[width=0.92\linewidth]{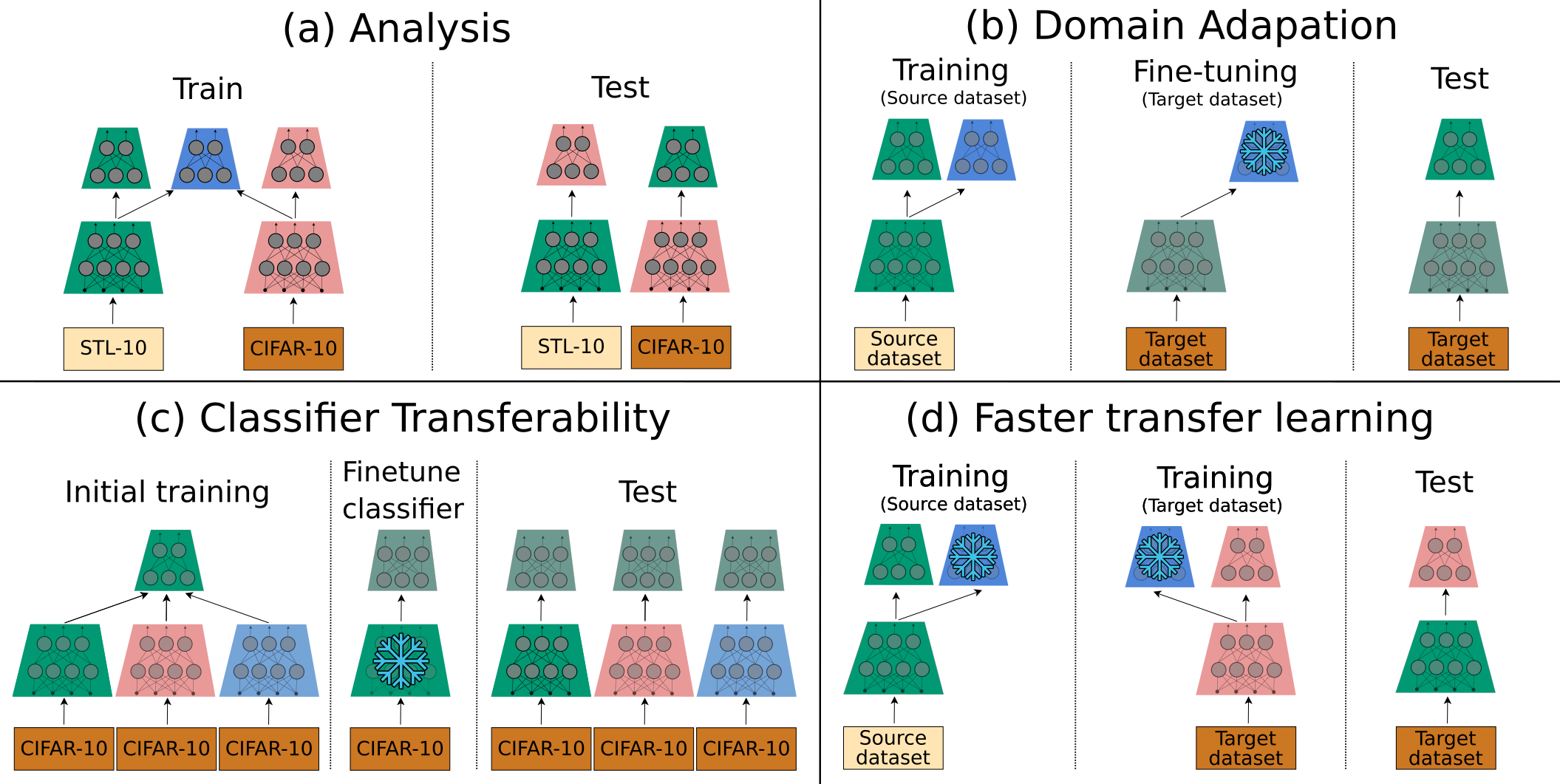}
\caption{
\textbf{Experimental setups for the analysis and the applications.} Our method enables recombining network components, which benefits domain adaptation, classifier transferability and efficient transfer learning.}
\label{fig:setups}
\end{figure*}
\section{Introduction}
\label{sec:introduction}

In computer vision we often train a different neural network for each task, where reuse of previously learnt knowledge typically remains limited to pre-training on ImageNet (ILSVRC-12)~\cite{russakovsky15ijcv}. However, human knowledge is composable and reusable~\cite{tenenbaum11science}.
Therefore it seems prudent to give neural networks these properties too.
Similar to what humans do, computer vision methods should reuse and transfer from previously acquired knowledge in the form of previously trained models~\cite{zamir18cvpr, ngiam18arxiv, dwivedi19cvpr, achille19iccv}.
For example, when a model can recognize cars in daylight, this knowledge should help recognizing cars by night through domain adaptation~\cite{ben10ml,shu18iclr}. 
In addition, when a model expands its knowledge,~\eg through more training examples or by learning a new concept, these improvements should be easily transferable to 
related tasks.

We believe that a general way to achieve network reusability is to build a large library of \emph{compatible components} which are specialized for different tasks. For example, some would extract features from RGB images, depth images, or optical flow fields. Other components could use these features to classify animals, localize cars, segment roads, or estimate human body poses.
The compatibility of the components would make it easy to mix and match them into a highly performing model for the task at hand.
Besides domain adaptation and transfer learning, this would also enable training a single classifier which can be deployed on various devices, each with its own hardware-specific backbone network.
We make a first step in the \new{direction of reusable components} by devising a training procedure to make the feature representations learnt on different tasks become compatible, without any post-hoc fine-tuning.
On the long term, we envisage a future where the practice of building computer vision models will mature into a state similar to the car manufacturing or building construction industries: with a large pool of high-quality and functionally well-defined compatible parts that a designer can conveniently recombine into more complex models tailored to new tasks.
The compatibility of components saves the designer the effort to make them work together in a new combination, so they are free to focus on designing ever more complex models.

Our quest for reusable components is related to the question of how similar the representations of independently trained networks are, when they are trained on similar data~\cite{kornblith19icml,lenc19ijcv,li16iclr,lu18nips,mehrer18ccn,morcos18nips,wang18nips}. 
Instead of such a post-hoc analysis, we make a first step towards training neural network that are \emph{directly compatible}, rather than only similar (\eg in terms of feature correlation~\cite{li16iclr,morcos18nips}).
For the purpose of this paper, we define components by splitting a neural network into two parts: a \featureextractor{} and a target \taskhead{}.
We say two networks are \emph{compatible} if we can recombine the \featureextractor{} of one network with the \taskhead{} of the other while still producing good predictions, directly without any fine-tuning after recombination (Fig.~\ref{fig:setups}).
When network components become perfectly compatible, they can be interchanged at no loss of accuracy.
It is important to note that compatibility does not require learning \emph{identical} mappings, as in feature distillation~\cite{romero15iclr}. 
As an extreme case, we could not distill the features from a network for street view images to a network for underwater images.
Instead, by only requiring compatibility as defined above, each network can learn a \featureextractor{} that is appropriate for its task.

Concretely, we introduce three ways to alter the training procedure of neural networks to encourage compatibility (Sec.~\ref{sec:method}):
using a shared self-supervised auxiliary head which predicts rotation~\cite{gidaris18iclr} (Sec~\ref{sec:ss_alignment}),
using a shared auxiliary head which discriminates common classes (Sec.~\ref{sec:class_alignment}),
and starting training from identical initial weights (Sec.~\ref{sec:iiw}).
We systematically analyse how well our methods make components compatible for the case of two image classification networks, one for CIFAR-10 and the other for STL-10 (Sec.~\ref{sec:analysis}).
We also demonstrate that compatibility comes at no loss of accuracy on the original tasks.
In Sec.~\ref{sec:applications} we apply our method to three diverse applications: unsupervised domain adaptation (Sec.~\ref{sec:uda}), transferring pre-trained classifiers across networks with different architectures (Sec.~\ref{sec:backbones}) and increasing the computational efficiency of transfer learning (Sec.~\ref{sec:faster_transfer}).
These applications involve demonstrating compatibility between networks with different architectures, making several networks compatible at the same time, and testing on more complex datasets like CIFAR-100 and ILSVRC-12.

\vspace{4pt}
\section{Related Work}
\label{sec:related_work}
\new{In this section we discuss the relation of our work with existing methods. We provide a structured positioning in Tab.\ref{tab:positioning}.}
~\begin{table*}[htb]
\centering
\footnotesize
\renewcommand{\arraystretch}{1}
\begin{tabular}{p{2.7cm}|p{2.2cm}|p{1.9cm}|p{1.9cm}|p{1.9cm}|p{2.3cm}|p{1.9cm}}
 & \textbf{Representational similarity} & \textbf{Distillation} & \textbf{Multi-task learning} & \textbf{Continual learning} & \textbf{Unsupervised domain adaptation} 		& \textbf{Compatibility}   \\
 \hline
Analysis / application 			& Analysis 				& Application	& Application	& Application	& Application 			& Both \\
Paired data 					& Yes 					& Yes 				& No			& No 			& No					& No \\
Identical output space			& No  					& No  			& No			& No			& Yes 					& No \\
Identical architecture 			& No  					& No			& Yes			& Yes			& Yes					& No \\
\new{Learns from / adapts existing model}		& Not applicable		& Yes			& No			& Yes			& Yes 					& No \\
Feature space alignment method & Post-hoc, Per-example & Per-example 	& Shared computational path & Various	& Distribtion matching & Compatibility \wrt{} a head\\
\end{tabular}
\caption{\new{\textbf{Positioning of compatibility \wrt related work.}}
}
\label{tab:positioning}
\end{table*}

\para{Representational similarity analysis.}
Several works investigate whether neural networks learn different projections of the same high-level representations, when trained independently and potentially on different datasets~\cite{li16iclr,lu18nips,mehrer18ccn,morcos18nips,wang18nips,lenc19ijcv,kornblith19icml,tang20arxiv}.
Closest to our work,~\cite{lenc19ijcv} analyze representational similarity via the performance of networks after recombining their components. As they start from independently trained networks, they require adding a stitching layer and training the recombined network with a supervised loss for several epochs.
We alter training, instead, so that the components are directly compatible and can be recombined, without the need for post-hoc optimization.
This leads to features that are more similar than those of independently trained networks, as we show using~\cite{li16iclr}.

\para{Distillation \& per-example feature alignment.}
Methods for feature alignment aim at training two or more networks so that they map a data sample to the same feature representation. 
Inspired by knowledge distillation~\cite{bucilua06kdd,hinton15arxiv}, feature distillation trains a network to approximate the feature activations of another network~\cite{romero15iclr}.
Instead, multi-modal embeddings methods learn to map multiple views of the same example to a common representation~\cite{frome13nips,socher13nips,karpathy15cvpr,gupta16cvpr,wang18cvpr}. 
Examples include methods for image captioning, which train on image+caption pairs~\cite{karpathy15cvpr}, or mix-and-match networks~\cite{wang18cvpr}, which map different modalities to the same representation,~\eg depth+RGB pairs.
In contrast,
our notion of compatibility does not require both networks to be trained from the same data, nor to see paired views of the same data. It is thus more generally applicable.

\para{Multi-Task Learning (MTL).}
The goal of MTL is producing {\em a single} network which can solve multiple tasks. This is typically achieved with large architectures and partially sharing computational paths across tasks,~\eg~\cite{misra16cvpr,kaiser17arxiv,maninis19cvpr}.
In contrast, we train {\em different} networks whose components are compatible and our method regularizes the feature representation space only, without imposing architectural constraints. 
Furthermore, MTL requires access to all datasets of all tasks at the same time for training, which becomes burdensome in computation and engineering as the number of task grows, and might be infeasible due to licensing or privacy concerns.
Instead, our incremental version (Sec.~\ref{sec:training_schemes}) does not require simultaneous access to all datasets, which we demonstrate experimentally in Sec.~\ref{sec:analysis}, \ref{sec:uda} \& \ref{sec:faster_transfer}.

\para{Continual learning (CL).}
CL is typically formulated as the online version of MTL~\cite{ruvolo13icml,rebuffi17cvpr,farquhar18arxiv,chen18aiml,parisi19nn}.
Thereby, the core challenge is to preserve compatibility between existing classification heads and a \featureextractor{} that gets updated over time. 
This is often addressed by penalizing changing important weights~\cite{kirkpatrick17pnas,zenke17icml,aljundi18eccv}, or changing predictions of the model on previous tasks~\cite{li2017learning,shmelkov17iccv,michieli19iccvw}.
Instead of just \emph{preserving} compatibility between initially identical networks, we propose a method that \emph{produces} compatible networks even if they have different architectures, are trained on different datasets, or start from different initializations (Sec.~\ref{sec:experiments} \&~\ref{sec:applications}).

\para{Unsupervised domain adaptation (UDA).}
The goal of UDA is to produce a model which works on the target domain, given a labeled source domain but only unlabeled data form the target domain.
There are two dominant ways to approach this~\cite{wang18neurocomputing,zhangw18cvpr}: 
(i) train one model that works on both the source and the target domain,
\eg~\cite{tzeng2014arxiv,ganin15icml,zhangw18cvpr,kumar18neurips},
who make features domain invariant through a domain discriminator; or
(ii) train a model on the source domain and then adapt it to the target domain,
\eg~\cite{saito17icml,zhangw18cvpr}, who rely on pseudo labels.
The latter is more general since it does not require access to the source and target dataset at the same time,
and also works for non-conservative domain adaptation where a single classifier cannot perform well in both
domains~\cite{ben10ml,shu18iclr}.
In Sec. 5.3 we adopt this approach and show that the self-supervised version of our method improves UDA.

\para{Transfer Learning (TL).}
The goal of TL is to improve results on a target task by reusing knowledge derived from a related source task. The current standard is to simply reuse the \featureextractor{} of a model trained on ILSVRC-12~\cite{donahue13decaf,sharif14cvprw,ren15nips,he2017mask}. However, what is the best source task depends on the target task~\cite{zamir18cvpr, ngiam18arxiv, dwivedi19cvpr, achille19iccv, yan20arxiv_b}. In~\cite{zamir18cvpr}, they proposed a computational framework to find good source tasks. But this is expensive:
finding good source+target task combinations consumed 50'000 GPU hours to train 3000 transfer functions.
Recently,~\cite{dwivedi19cvpr,achille19iccv} proposed methods to predict what tasks to transfer from.
This allows to only transfer, fine-tune, and test the most promising \featureextractor{}s, thus saving computation.
In Sec.~\ref{sec:faster_transfer} we demonstrate that we can reduce the amount of fine-tuning necessary to achieve good performance on the target task, which increases efficiency further.

\vspace{4pt}
\section{Method}
\label{sec:method}
We consider neural networks formed by the combination of two components: A \emph{feature extractor} $f(\cdot)$ and a \emph{target task head} $h(\cdot)$, parameterized by $\bPhi$ and $\bTheta$, respectively. In standard supervised learning, one trains a neural network on task $t$ by minimizing a task loss
$\ell_t(h(f(\bx_i; \bPhi_t); \bTheta_t), \by_i)$ 
over all examples $\bx_i$ with label $\by_i$ in dataset $\dataset_t$.
We denote a standard network trained on task $t$, using the feature extractor and target head of task $t$, as $n_{tt}(\bx_i)$.

When independently training two networks on tasks $a$ and $b$ by minimizing their respective losses $\ell_a(\cdot)$ and $\ell_b(\cdot)$, the resulting networks are incompatible: 
Recombining their components into a new network $n_{ab}(\bx_i) = h(f(\bx_i; \bPhi_a); \bTheta_b)$
or $n_{ba}(\cdot)$
produces random or systematically wrong predictions (Sec.~\ref{sec:experiments}).
This happens because the two feature extractors generally learn features responding to different image patterns, with different scaling of activation values, and even equivalent feature channels will appear in arbitrary orders~\cite{lenc19ijcv,kornblith19icml}.

\para{Compatibility.}
Our goal is to achieve compatibility between networks, directly after training.
We define compatibility based on the performance of the recombined networks $n_{ab}(\cdot)$ and $n_{ba}(\cdot)$.
When these network performs at chance level, we say that the components of $n_{aa}(\cdot)$ and $n_{bb}(\cdot)$ are {\em incompatible}.
Instead, they are \emph{compatible} when $n_{ab}(\cdot)$ and $n_{ba}(\cdot)$ directly output predictions that are significantly better than chance, without any fine-tuning after recombination.
Generally, the recombined networks will not exceed the performance of the vanilla networks $n_{aa}(\cdot)$ and $n_{bb}(\cdot)$, trained and tested on their own task without recombining any component.
Thus, we define this performance as the practical upper bound. When the recombined networks reach this upper bound, they are \emph{perfectly compatible}, which allows to use their components interchangeably.

To achieve compatibility, we introduce some degree of dependency between the training processes of $n_{aa}(\cdot)$ and $n_{bb}(\cdot)$.
Specifically, we encourage compatibility between the features produced by their extractors $f(\bx_i;\bPhi_a)$ and $f(\bx_i;\bPhi_b)$.
As many different parameterizations of a neural network produce comparable performance~\cite{choromanska15ais,lu18nips}, we
hypothesize that we can make networks more compatible without decreasing the performance on their original task (confirmed in our experiments in Sec.~\ref{sec:analysis}).

Next, we introduce three different methods that encourage compatibility.
For clarity of exposition, we describe the case for two networks, but our methods works with any number of networks (Sec.~\ref{sec:backbones} \&~\ref{sec:faster_transfer}).
Similarly, while we denote the model components with $f(\cdot)$ and $h(\cdot)$ for simplicity, our method also handles the case where networks $n_{aa}$ and $n_{bb}$ have a different architecture (Sec.~\ref{sec:backbones}).

\subsection{Compatibility through Self-Supervision (\RP{})}
\label{sec:ss_alignment}

We propose to make components compatible via a
generally applicable \auxiliarytask{}, based on a self-supervised objective.
Self-supervision relies on supervised learning techniques, but the labels are created from the unlabelled input data itself.
We adopt the approach of previous methods like~\cite{noroozi16eccv,doersch17iccv,gidaris18iclr}.
First, we transform an image $\bx$ with $g\left(\bx,\bs\right)$, a function which applies a transformation $\bs$.
Then, the task of the network is to predict what transformation was applied (its label).

To achieve compatibility, this \auxiliarytask{} has its own head $s$, but operates on the features produced by the extractors of the respective target tasks (Fig.~\ref{fig:setups}a). Specifically, its prediction function is $s(f(\bx;\bPhi_t);\bTheta_{s})$, where $\bTheta_s$ are the parameters of the \auxiliarytaskhead{}.
During training, we minimize the \targettask{} losses and the \auxiliarytask{} loss for both tasks:
\begin{equation}
\begin{split}
    \sum_{{t \in \{a,b\}}}
    \sum_{\substack{(\bx_i, \by_i) \in \dataset_t}}
    \Big[    
\!\ell_t \left(h\left(f\left(\bx_i; \bPhi_t\right); \bTheta_t\right), \by_i\right) \\
+ \! \frac{1}{|\mathcal{S}|} \sum_{{\bs \in \mathcal{S}}}
\ell_s \left(h\left(f\left(g\left(\bx_i,\bs\right); \bPhi_t\right); \bTheta_s\right), \bs\right)
\Big] 	
\end{split}
\label{eq:rp}
\end{equation}
where $\mathcal{S}$ is set of possible transformations that are applied, $\bTheta_s$ are the parameters of the auxiliary task head, and $\ell_s$ its associated loss.
While there are \targettask{} parameters $\bPhi_t$ and $\bTheta_t$ specific to each task, we tie the \auxiliarytask{} parameters $\bTheta_{s}$ across tasks.
This forces the feature extractors $f(\bx_i;\bPhi_t)$ of each task $t$ to produce features that are compatible with the same \auxiliarytaskhead{}.
As we show in Sec.~\ref{sec:analysis}, this leads to feature extractors that are compatible more generally, allowing to recombine the feature extractor of one with the \targettaskhead{} of the other.

\para{Choice of self-supervision task.}
Throughout this work we use rotation prediction~\cite{gidaris18iclr}.
The input image is transformed by rotating it with an angle $\mathcal{S} = \{0\degree,90\degree,180\degree,270\degree\}$ and the task is to classify which rotation angle was applied.
For simplicity we refer to this method as \textit{compatibility through \rp} (RP), but any other self-supervised objective can be used here.
We discuss considerations for choosing a suitable self-supervised task in 
Appendix~\ref{sec:additional_discussion}.

\para{Trade-offs.}
This compatibility method is very general. It only requires the shared self-supervised task to be both meaningful and non-trivial~\cite{sun19arxiv_a,tschannen19arxiv}.
While such a task can be defined on almost any dataset, the quality of the induced compatibility depends on how much the \targettask{} and the \auxiliarytask{} rely on the same features.
In theory, a weakly related or orthogonal self-supervised \auxiliarytask{} could negatively affect the performance of the network on the \targettask{}.
\new{In practice though, it often improves performance~\cite{zhai19iccv,henaff19arxiv}.
Similarly, in our experiments we only observe positive effects on performance when adding \rp{}.}

\subsection{Compatibility through Discriminating Common Classes (\DCC{})}
\label{sec:class_alignment}
When tasks $a,b$ have common classes, we can directly use these to achieve compatibility, rather than resorting to a self-supervised loss. Hence, we propose an \auxiliarytaskhead{} $c$, which discriminates among these common classes.
Specifically, we minimize the following loss:
\begin{equation}
\begin{split}
    \sum_{{t \in \{a,b\}}}
    \sum_{\substack{(\bx_i, \by_i) \in \dataset_t}}
    \Big[
\ell_t \left(h\left(f\left(\bx_i; \bPhi_t\right); \bTheta_t\right), \by_i\right) \\
 + \! \ell_c \left(h\left(f\left(\bx_i; \bPhi_t\right); \bTheta_c\right), \by_i\right)
\cdot\indicator\left[\by_i\!\in\!\mathcal{C}\right]
\Big]
\end{split}
\label{eq:dcc}
\end{equation}
where $\ell_c$ is the \auxiliarytask{} loss. It is computed only over examples in the set of common classes $\mathcal{C}$ ($\indicator$ is an indicator function returning 1 if its argument is true and 0 otherwise).

\para{Trade-offs.}
While this method is expected to achieve high compatibility, it requires the target tasks to have common classes. Depending on the scenario, the target tasks might actually have few or even no common classes.

\subsection{Compatibility through Identical Initial Weights (\IIW{})}
\label{sec:iiw}

\cite{zhang19icmlw} demonstrated that for many layers in a trained network, resetting the weights of that layer to their initial values leads to a limited loss in accuracy.
This suggests that the initialization defines a set of random projections which strongly shape the trained feature space. Hence, we propose to encourage compatibility simply by starting the loss minimization of both tasks from \iiw{} (\IIW).
For this method, we initialize using either identical \emph{random} weights or identical \emph{pre-trained} weights (Sec.~\ref{sec:analysis}).

\para{Trade-offs.}
This method only works when both tasks have identical network architectures.
Moreover, it only acts at the start of training, where it makes networks identical and thus perfectly compatible.

\subsection{Training schemes}
\label{sec:training_schemes}
For \RP{} and \DCC{} we consider two training schemes: \emph{\jointtraining{}} and \emph{\incrementaltraining{}}.

In \emph{\jointtraining{}}, we minimize~\eqref{eq:rp} (or~\eqref{eq:dcc}) by alternating between tasks $a$ and $b$, each time minimizing the loss over a single minibatch. This resembles multi-task training~\cite{doersch17iccv,maninis19cvpr},
but here each task has its own network,
rather than having a single network with shared computation.
By training jointly, both target tasks ${a, b}$ influence the \auxiliarytaskhead{} parameters and use that head to solve the \auxiliarytask{}.

In \emph{\incrementaltraining{}}, we first train the network $n_{aa}$ by minimizing~\eqref{eq:rp} (or~\eqref{eq:dcc}) over task $a$ only.
This also learns the parameters of the \auxiliarytaskhead{}.
Later, we train the network $n_{bb}$ on task $b$, but use the \auxiliarytaskhead{} with its parameters frozen.
This encourages compatibility between $n_{aa}$ and $n_{bb}$, without requiring both of them to be trained at the same time.

\vspace{4pt}
\section{Analysis of compatibility}
\label{sec:experiments}
\label{sec:analysis}
\begin{figure*}[t]
\centering
\includegraphics[clip,trim=0 10pt 0 0,width=1\linewidth]{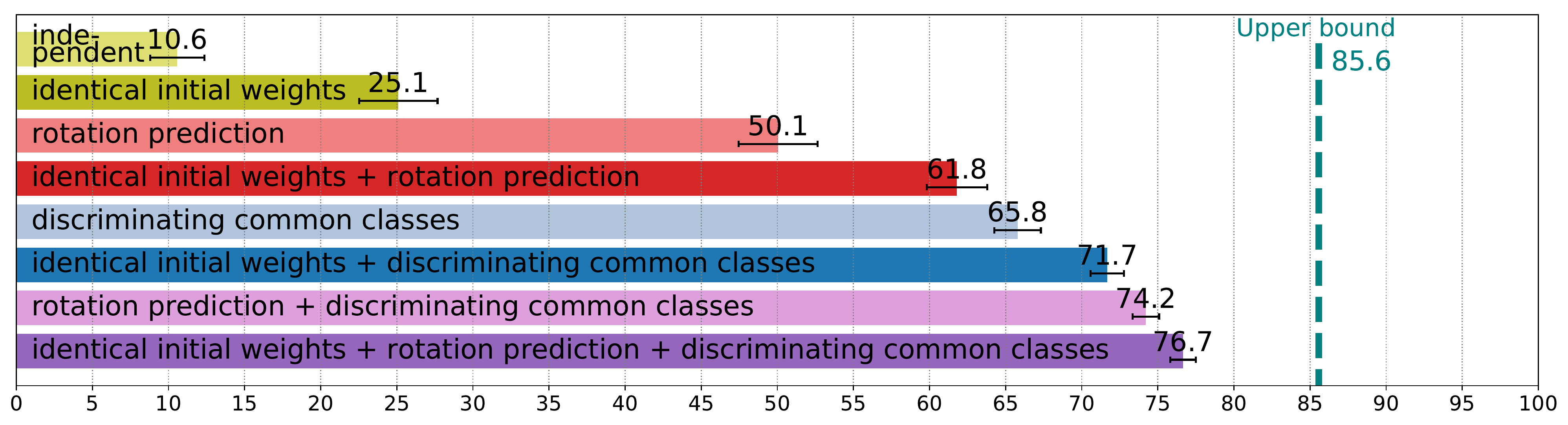}
\caption{
\small{
  \textbf{Recombination accuracy for different methods.}
We report \accuracy{} with standard deviations over 10 runs (horizontal line segments).
Results for each dataset seperately are given in
Appendix~\ref{sec:per_dataset_results}.
}
}
\label{fig:method_comparison}
\end{figure*}
We now analyse the induced compatibility of our methods introduced in Sec.~\ref{sec:method}: \dcc{} (\DCC), \rp{} (\RP), and initializing networks with \iiw{} (\IIW).

\para{Experimental setup.}
Fig.~\ref{fig:setups}a illustrates our basic experimental setup when using a single auxiliary task (\DCC{} or \RP{}).
As network architecture we use ResNet-56~\cite{he16cvpr}, which consists of 3 stages with 9 ResNet blocks of two layer each. We split this into a feature extractor and target task head directly after the second stage (results for other splits are in Appendix~\ref{sec:layer_compatibility}).
We train one network on the CIFAR-10~\cite{krizhevsky09} train set and one on the STL-10~\cite{coates11aistats} train set. These datasets have 9 classes in common.
For simplicity we mostly train networks jointly in this analysis. We also briefly explore incremental training (Sec.~\ref{sec:training_schemes}) which we use extensively in Sec.~\ref{sec:applications}.

An important detail is that our network components use Batch Normalization (BN)~\cite{ioffe15icml}. At training time, BN normalizes the features in each batch to have zero mean and unit variance.
At test time, features are normalized using aggregated statistics over the train set.
However, \DCC{} and \RP{} encourage compatibility in the training regime of single-batch statistics, which may vary wildly per task. This makes the aggregated training statistics unreliable for any recombination of components.
Therefore we use batch statistics at test time in all experiments (see
Appendix~\ref{sec:bn} 
for more discussion \new{and alternatives}).

As metric we define {\em \accuracy{}}:
We recombine the CIFAR-10 feature extractor with the STL-10 classification head and measure accuracy on CIFAR-10 test,
on the 9 common classes. We measure accuracy immediately after recombination, without any fine-tuning. We do the analogue for STL-10 and report the average over the two test sets.

\para{Evaluation of methods to encourage compatibility.}
Fig.~\ref{fig:method_comparison} shows recombination accuracy for our different compatibility measures.
While the independently trained networks perform at chance level (10.6\%), our proposed methods achieve good levels of compatibility: \DCC{} works best (65.8\%), followed by \RP{} (50.1\%).
We note that \RP{} is more generally applicable, since it does not not require any common classes.
We investigate \accuracy{} as a function of the number of common classes in
Appendix~\ref{sec:dcc_varying_classes}.

Interestingly, there is even some compatibility between networks just by starting from \iiw{} and then separately minimizing the task loss on the two different datasets (\IIW{}: 25.1\%).
The experiments also show that all three methods are complementary: using all methods together reaches 76.7\% \accuracy{}.

As upper bound, we use the classical setting of training and testing a network on each task separately. For fairness, we give each network its own rotation prediction head \new{which we found to improve results by 2.0\%, but which} does not encourage compatibility. This results in an upper bound of 85.6\% accuracy. Given that the tasks are different, we consider \IIW+\RP+\DCC{} (at 76.7\%) to come rather close to this upper bound.

Importantly, we achieve compatibility without compromising accuracy on the original tasks.
Using \IIW+\RP+\DCC{} on the original networks and measuring accuracy on their own target tasks without recombination reaches 86.3\%. This is slightly higher than our upper bound, likely because of beneficial regularization.

\para{Starting from pre-trained models.}
It is common to start from a model pre-trained for ILSVRC-12 classification~\cite{donahue13decaf,ren15nips,he2017mask}.
Therefore we repeat the above experiments but starting from models pre-trained for self-supervised rotation prediction on ILSVRC-12
(details in Appendix~\ref{sec:implementation_details}).

For our \IIW{} experiments we initialize both networks using the same pre-trained weights.
When not using \IIW{}, we initialize the two networks with \emph{different} pre-trained weights.

Initializing networks using the same pre-trained weights (\IIW{}) strongly encourages compatibility and already leads to a \accuracy{} of 74.3\%. The strongest compatibility is achieved by combining \IIW{} with \DCC{} (82.7\%).

Experiments not using \IIW{} exhibit a counter-intuitive effect. For \RP{} we reported 50.1\% \accuracy{} when initializing the two networks using different \emph{random} weights (Fig.~\ref{fig:method_comparison}). Now, when initializing using different \emph{pre-trained} weights, \accuracy{} drops to 19.7\%. Similarly, for \DCC{} \accuracy{} drops from 65.8\% to 52.3\%.
This suggests it is \emph{harder} to make networks compatible after they are already independently (pre-)trained. Compatibility should thus be encouraged from the beginning of the training process.

\para{Joint \vs incremental training.}
While so far we trained the two networks jointly, some practical applications require making a network compatible with an existing one (Sec.~\ref{sec:applications}).
Therefore we analyze here what happens when we train networks \emph{incrementally} (Sec.~\ref{sec:training_schemes}). We first train a network on CIFAR-10 with auxiliary task heads \DCC{} and \RP{}. We then freeze the \DCC{} and \RP{} heads. Finally, we train a new network on STL-10 starting from identical initial weights (\IIW{}) and also using the frozen \DCC{} and \RP{} heads. We repeat the analogue starting from STL-10. 

Compared to joint training in Fig.~\ref{fig:method_comparison}, results decrease moderately from 76.7\% to 72.7\%. This demonstrates that we can make new networks compatible with existing ones.

\para{Reaching the compatibility upper bound.}
While we achieved high compatibility in our experiments involving CIFAR-10 and STL-10 (Fig.~\ref{fig:method_comparison}), we did not reach the upper bound and hence our components are not \emph{perfectly compatible}.
However, a classification head optimized for CIFAR-10 is not expected to yield top accuracy on STL-10 (and vice versa).
To remove the task mismatch, we repeat all experiments using CIFAR-10 for both tasks.
In this setting, several combinations of methods reach the upper bound:
\IIW+\RP, \IIW+\DCC, and \IIW+\RP+\DCC{} (see
Appendix~\ref{sec:upper_bound}).
This shows that our methods are in principle strong enough to achieve perfect compatibility, when the data allows it.

\begin{figure*}[t]
\centering
\includegraphics[width=1\linewidth]{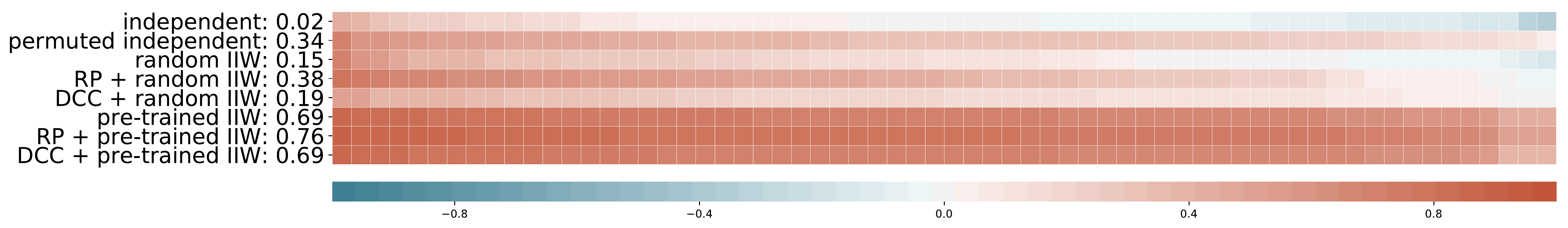}
\caption{
\textbf{Cross-Correlation between features produced by two feature extractors.} The blocks visualise the per-channel cross-correlations, sorted by magnitude. The number reports the cross-correlation averaged over all channels.}
\label{fig:feature_similarity}
\end{figure*}

\para{Feature cross-correlation.}
\new{Various analysis papers measure how similar, under some transformation, the features produced by two networks are.
Their goal is to understand whether the networks activate on similar image patterns and hence learn similar representations (discussed in Sec.~\ref{sec:related_work}). Following~\cite{li16iclr}, we repeat this analysis for two independently trained networks. We feed all images from the full CIFAR-10 and STL-10 test sets to both feature extractors and measure cross-correlation under an optimal permutation, as determined post-hoc (denoted as 'permuted independent'). For our methods of compatibility, instead, we measure cross-correlation of the features produced by two networks directly after training,~\ie \emph{without} permutation, as these methods aim to directly make features compatible. Results are shown in Fig.~\ref{fig:feature_similarity}.
}

\new{
We observe that all our compatibility methods directly yields reasonably correlated features ($\geq 0.15$). While correlation for the post-hoc aligned features is $0.34$ (permuted independent), we measure higher correlation for \RP + random \IIW ($0.38$) and very high correlation when starting from pre-trained \IIW: ($0.69-0.76$). Hence, using pre-trained weights leads to strongly correlated features.
However, we also find that \RP{} leads to more strongly correlated features ($0.38$) than \DCC{} ($0.19$), even though \DCC{} leads to a higher \accuracy{} (Fig.~\ref{fig:method_comparison}).
Similarly, when using pre-trained weights, \IIW{} yields the same correlation as \IIW{}+\DCC{}, yet the latter yields 8.5\% higher \accuracy{} (Fig.~\ref{fig:method_comparison}).
We therefore conclude that a higher cross-correlation does not necessarily translate to more compatible features as measured by \accuracy{}.
Hence trying to learn identical feature mappings (as in feature distillation, \eg~\cite{romero15iclr}) is not required for compatibility.
}

\vspace{4pt}
\section{Applications}
\label{sec:applications}

\subsection{Unsupervised domain adaptation}
\label{sec:uda}

\para{Application.}
We transfer knowledge from a source domain with labeled data to a target domain with unlabeled data.

\para{Experimental setup (Fig.~\ref{fig:setups}b).}
\new{We first train a model on the source training set, where the model consists of a} \featureextractor{}, a \classificationhead{} and an auxiliary rotation prediction head \RP{} (initialized by training for \rp{} on ILSVRC-12~\cite{russakovsky15ijcv}).
We then want to adapt the \featureextractor{} of this source model to the target domain while preserving compatibility with the original classification head. 
We do this by freezing the \RP{} head while fine-tuning the \featureextractor{} on the unlabeled target training set. For this we minimize the self-supervised \RP{} loss for \att{1000} steps.
Finally, we recombine this updated \featureextractor{} with the source domain \classificationhead{} to predict classes on the target domain. We report average class accuracy on the target test set.
We evaluate adapting between CIFAR-10 and STL-10, as is common in this area~\cite{ghifary16eccv,shu18iclr,sun19arxiv_a}.
We use here a larger WRN-28~\cite{zagoruyko16bmvc} architecture as in~\cite{sun19arxiv_a} (see Appendix~\ref{sec:implementation_details}).

\para{Results (Tab.~\ref{tab:uda}).}
We compare our method to previous approaches and two baselines based on our source model.
One baseline uses the model as is. The other updates BN statistics at test time, which performs significantly better. This confirms their importance, as discussed in Sec.~\ref{sec:analysis} and observed by~\cite{li16icrlw}.
Our method improves performance further and matches the state-of-the-art on adapting from CIFAR-10 to STL-10~\cite{lee19iccv} (\att{82.6\%}). The methods \cite{kumar18neurips,sun19arxiv_a} perform best for adapting from STL-10 to CIFAR-10. On average over both adaptation directions, our method is competitive (\att{78.8\%} for~\cite{kumar18neurips} \vs \att{77.9\%} for us).

Importantly, our method is simpler and faster than competing methods. %
The state-of-the-art~\cite{kumar18neurips} combines multiple models, includes a domain discriminator~\cite{ganin15icml,ganin16jmlr}, employs a custom network architecture~\cite{shu18iclr}, and trains for 80000 steps on the joint source and target training sets.
Instead, we use a single ResNet model and fine-tune only for 1000 steps on the target domain, which makes our method computationally faster.
Finally,~\cite{sun19arxiv_a} gets significant gains by combining multiple self-supervised objectives, which we could potentially include as well.

\begin{table}
\footnotesize
\setlength\tabcolsep{2pt}
\begin{tabular}{p{2cm}|lcccHHH}
\hline
\multirow{2}{*}{\textbf{Method}} & Source: 										& CIFAR-10 	& STL-10 	& \multirow{2}{*}{Avg.}		& ImageNet-60	& CIFAR-60 	& \multirow{2}{*}{Avg.}		\\
& Target:							 			& STL-10 	& CIFAR-10	& 							& CIFAR-60		& ImageNet-60 & \\
\hline
\multicolumn{2}{l}{VADA \cite{shu18iclr}}  						& 78.3\%		& 71.4\%  	& 74.9\%						&				&		\\
\multicolumn{2}{l}{VADA+Co-DA$^{bn}$~\cite{kumar18neurips}}			& 81.3\%		& 76.3\%		& 78.8\% & & \\
\multicolumn{2}{l}{DTA~\cite{lee19iccv}}						& 82.6\%		& 72.8\% 		& 77.7\% & & \\
\hline
\multicolumn{2}{l}{Joint w/ rotation \cite{sun19arxiv_a}} 		& 81.2\%		& 65.6\%		& 73.4\%						&		\\
\multicolumn{2}{l}{Joint multi-objective \cite{sun19arxiv_a}} & 82.1\%		& 74.0\%  	& 78.1\%						&		\\
\hline
\multicolumn{2}{l}{Our source model} & 77.2\% 		& 52.9\%		& 65.1\%						& 				&  \\
\multicolumn{2}{l}{Our source model w/ test BN} & 82.0\%		& 71.3\%		& 76.7\%					& 				&  \\
\multicolumn{2}{l}{Ours w/ adaptation through \RP{}} & 82.6\%		& 73.1\%		& 77.9\%					& 				&  \\
\hline
\end{tabular}
\caption{
  \textbf{Accuracy for unsupervised domain adaptation.}
} 
\label{tab:uda}
\end{table}

\subsection{Compatibility across feature extractors with different architectures}
\begin{figure}[t]
     \includegraphics[width=1\linewidth,trim=0 5pt 10pt 0, clip]{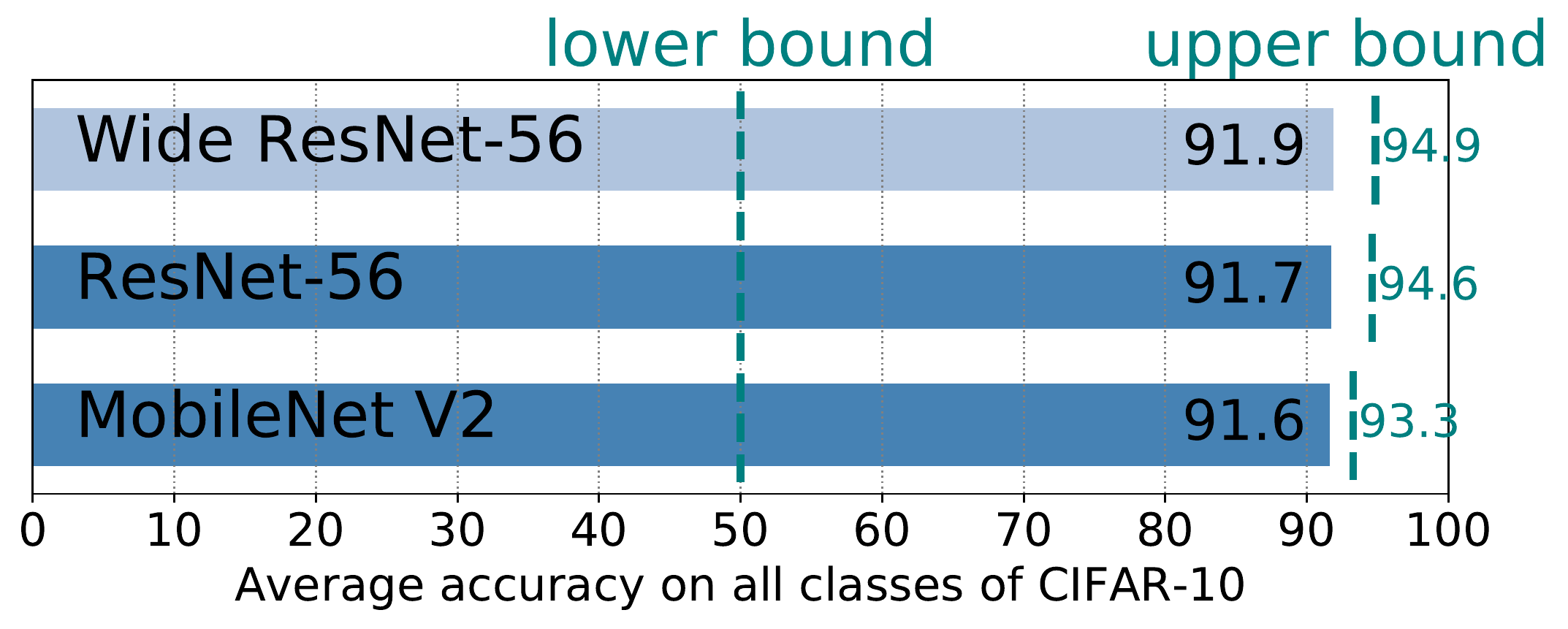}            
     \caption{\textbf{Accuracy when transferring a classification head to compatible feature extractors}.
     }
     \label{fig:multiple_backbones}
\end{figure}

\label{sec:backbones}
\begin{figure*}[t]
    \centering
     \includegraphics[trim={0cm 0cm 0cm 0cm},clip,width=0.96\linewidth]{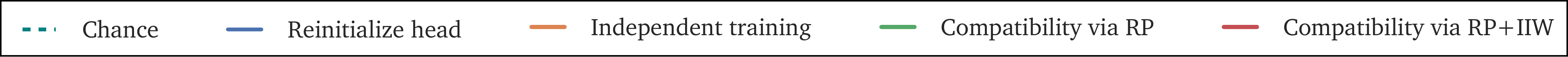}            
   \begin{subfigure}[t]{0.3\textwidth}
    \includegraphics[trim={0cm 0cm 0cm 0cm},clip,width=1\linewidth]{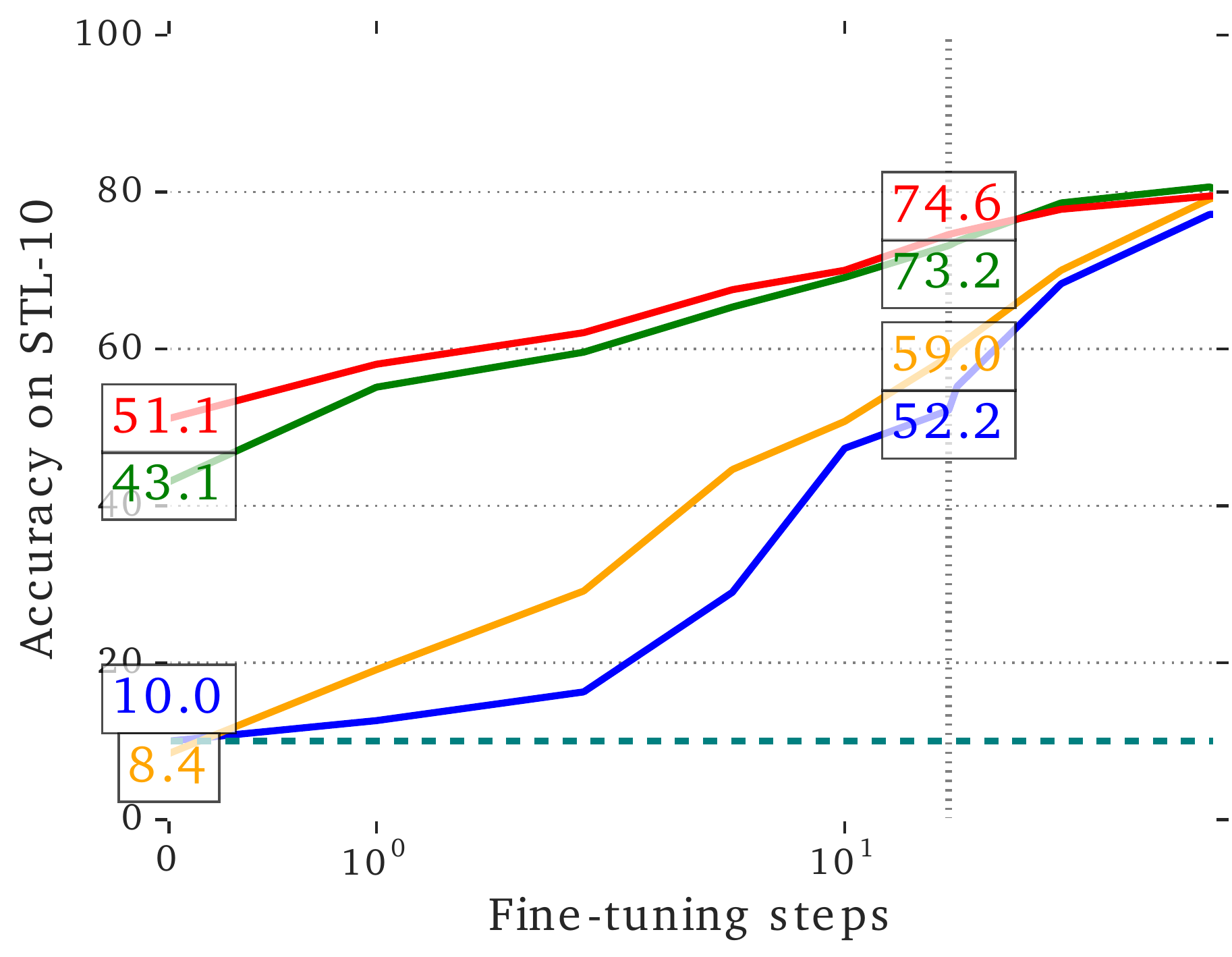}
    \caption{STL-10}
    \end{subfigure}
    \hspace{8pt}
       \begin{subfigure}[t]{0.3\textwidth}
    \includegraphics[trim={0cm 0cm 0cm 0cm},clip,width=1\linewidth]{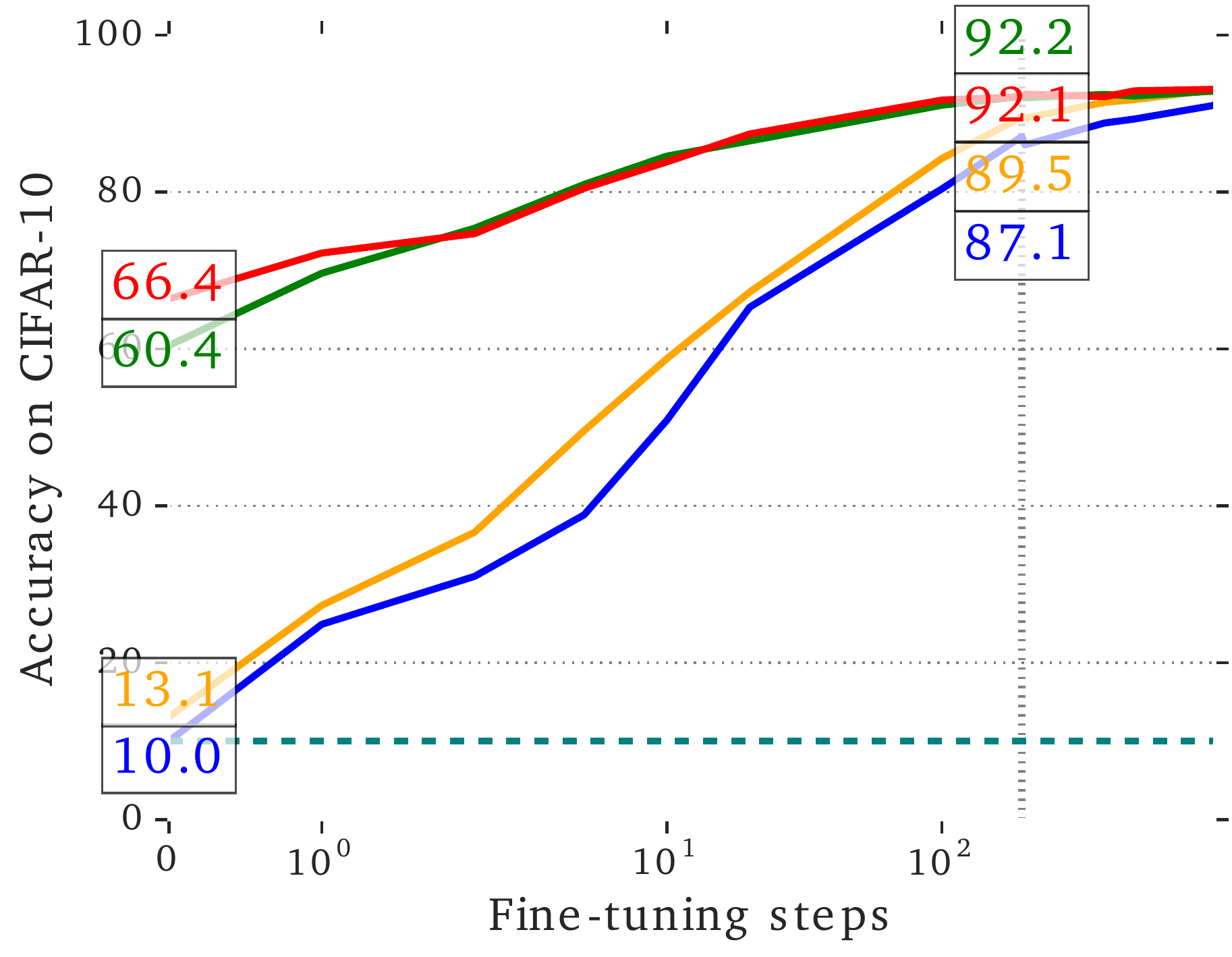}
    \caption{CIFAR-10}
    \end{subfigure}    
    \hspace{8pt}    
   \begin{subfigure}[t]{0.3\textwidth}
    \includegraphics[trim={0cm 0cm 0cm 0cm},clip,width=1\linewidth]{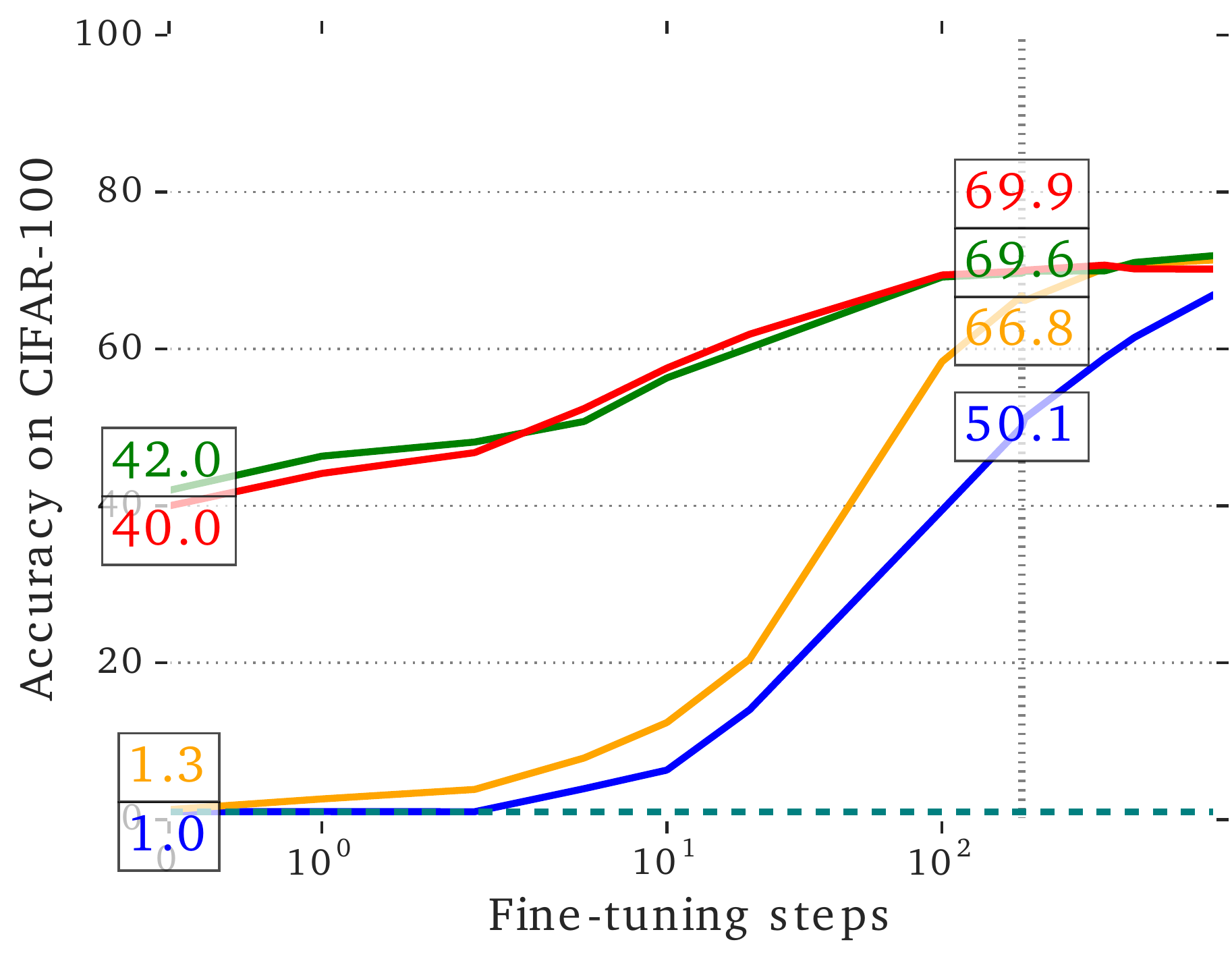}
    \caption{CIFAR-100}	
    \end{subfigure}        
    \caption{
\textbf{Average class accuracy of recombined components as a function of fine-tuning steps (log scale, up to 5 epochs).} 
We overlay accuracy values directly after recombination and after 1 epoch. Through training with \RP{}, components are compatible, enabling direct recombination.  
    }
    \label{fig:xp5}
\end{figure*}
\para{Application.} 
We want to achieve compatibility between feature extractors having different architectures, thus enabling transferring task heads across them. As a practical application we consider a single classification task which runs on many devices, each with a hardware-tailored network architecture (e.g. a powerful server, a standard desktop, a mobile phone). 
Normally, every time the set of classes to be recognized changes, all networks need to be retrained.
Instead, if their feature extractors are compatible, only one extractor and its corresponding classification head need to be retrained. We can then transfer that classification head to all other models. This greatly facilitates deployment of the updated classifier to all client devices, especially if different people are responsible for maintaining the different models.

\para{Experimental setup (Fig.~\ref{fig:setups}c).}
We consider three feature extractor architectures: ResNet-56~\cite{he16cvpr}, Wide ResNet-56~\cite{zagoruyko16bmvc}, MobileNet V2~\cite{sandler18cvpr}.
We combine these with a \DCC{} head based on MobileNet V2. In this application, \DCC{} not only encourages compatibility but also directly solves the target task (as there is just one task).
We split MobileNet V2 into components after the 11-th inverted ResNet block (out of 17).
To fit all extractors to a single \DCC{} head, we add to each extractor a 1x1 convolution layer with 64 output channels.
Differences in spatial resolution are resolved by the average pooling in the penultimate layer of the MobileNet V2 \DCC{}.

At first, we assume that we only have data for the first 5 classes of CIFAR-10. We use these to jointly train the three feature extractors with the \DCC{} head. At this point, each `feature extractor plus \DCC{}' network addresses the target task for a particular device.
Next, suppose we obtain labeled data for 5 new classes (resulting in the full CIFAR-10 training set). Instead of re-training everything, we only want to update the \DCC{} head. 
To do so, we first extend the classification layer of the \DCC{} head to handle 10 classes. Then, we choose the trained Wide ResNet-56 as the \emph{reference feature extractor}. We freeze it, attach the \DCC{} head, and fine-tune this combination on the CIFAR-10 training set. Finally, we attach the updated \DCC{} to each individual extractor and evaluate on the CIFAR-10 test set. Note that in this process we updated none of the feature extractors after the initial training phase (updating it is investigated in Appendix~\ref{sec:additional_classifier_transfer}).

\para{Results (Fig.~\ref{fig:multiple_backbones}).}
As an upper bound we train the individual networks on CIFAR-10 (also with a rotation prediction head). As an optimistic lower bound we consider perfectly discriminating the first five classes, leading to 50\% accuracy.
As Fig.~\ref{fig:multiple_backbones} shows, recombining either ResNet-56 or MobileNet V2 with the updated \DCC{} head lead to excellent accuracy of 91.7\% and 91.6\% respectively. While the upper bounds are even higher at 94.6\% and 93.3\%, our method requires much less computation and greatly facilitates deployment.
Part of the gap to the upper bound can be attributed to changes in architectures: we added 1x1 convolutions and use mixed architectures with a simple MobileNet head. If we redo the upper bound using these changed architectures, we get accuracies between 92.7\% and 92.8\%. This suggests that optimizing architectures would lead to even better results.

\subsection{Faster transfer learning}
\label{sec:faster_transfer}

\para{Application.}
In transfer learning, the goal is to improve results on a target task by reusing knowledge derived from a related source task.
In the deep learning era, the standard approach is to reuse the \featureextractor{} of a model trained on the source training set.
This source \featureextractor{} and a randomly initialized task-specific head are combined into a new model, which is then fine-tuned on the target training set.
When there are many possible source tasks,
this process is computationally expensive,\new{~\eg~\cite{zamir18cvpr} reports consuming 50'000 GPU hours.}
Instead, we propose to train an initial \targettaskhead{} and \emph{reuse} it when exploring different source tasks to transfer from (Fig.~\ref{fig:setups}d).
For this, we recombine the source \featureextractor{} and the initial \targettaskhead{} into a new model.
When these components are compatible, the benefits of transferring from a potential source can be evaluated and capitalized on with no or little fine-tuning on the target training set.

\para{Experimental setup (Fig.~\ref{fig:setups}d).}
We study transferring from a model trained on ILSVRC-12 as the source task. For this, we simply replace the \featureextractor{} of the \targettask{} model with the source one, while keeping the \targettask{} head.
We make these components compatible by training with \rp{} (\RP{}) and an incremental training scheme (Sec.~\ref{sec:training_schemes}).
In this scheme, we first need to set the weights of \RP{} ($\bTheta_s$), which we obtain by training a model on CIFAR-100~\cite{krizhevsky09}.
Then, the source and target models are trained with this frozen \rp{} head, forcing their \featureextractor{} to produce features that work with that same \rp{} head.

We compare our method against re-initializing the target task head or recombining independently trained components. For these baselines, we also use rotation prediction as an auxiliary task for fair comparison, but initialize the weights of its head randomly for each network.
We evaluate transferring a \featureextractor{} trained on ILSVRC-12~\cite{russakovsky15ijcv} to different target tasks, here CIFAR-10~\cite{krizhevsky09}, STL-10~\cite{coates11aistats}, or CIFAR-100~\cite{krizhevsky09}.
We measure transfer efficiency as the accuracy directly after recombination, and after a few epochs of fine-tuning on the \targettask{}.

\para{Results (Fig~\ref{fig:xp5}).}
Our method achieves strong results in terms of accuracy on the target task, even without any fine-tuning.
Here, the networks are trained separately and only made compatible via \RP{} and optionally \IIW{}.
Nonetheless, our method achieves 40.0\%-66.4\% \accuracy{}, despite the differences in the datasets and their class vocabularies.
Instead, the baselines yield random performance before fine-tuning, as expected.
After 1 epoch of fine-tuning our methods are still significantly better than the baselines.
They converge only after fine-tuning for several epochs (Fig~\ref{fig:xp5}).

In summary, our method reduces the need for fine-tuning when transferring components. As this is a core part of existing transfer learning methods
~\cite{zamir18cvpr, dwivedi19cvpr, achille19iccv, yan20arxiv_b},
our method can help speed these up.

\vspace{4pt}
\section{Conclusion}
\label{sec:conclusion}
We have demonstrated that we can train networks to produce compatible features, without compromising accuracy on the original tasks. We can do this through joint training, or by making new networks compatible with existing ones, through iterative training. By addressing three different applications, we demonstrated that our approach is widely applicable. 

{\small
\bibliographystyle{aaai21}
\bibliography{shortstrings,loco}
}

\iftrue
\appendix
\section{Choice of self-supervision task}\label{sec:additional_discussion}
As discussed in Sec. 3.1 of the main paper, we use rotation prediction~\cite{gidaris18iclr} as a self-supervised \auxiliarytask{}. We selected this task due to its simplicity and since it has been shown to work well for feature learning~\cite{gidaris18iclr,zhai19iccv,sun19arxiv_a,zhai19arxiv}.
Our method could however also be used with other standard self-supervised tasks, such as  solving jigsaw puzzles~\cite{noroozi16eccv}, colorization~\cite{zhang2016colorful} or exemplar classification~\cite{dosovitskiy14nips}. 
Importantly, self-supervised learning is continuously improved~\cite{he19arxiv, chen20arxiv, yan20arxiv,jing20pami}, where contrastive prediction has recently gained popularity~\cite{hadsell06cvpr,dosovitskiy14nips,oord18arxiv,bachman19nips,he19arxiv,chen20arxiv}. We hypothesize that these improvements will also translate to stronger compatibility when adopted in our method.

Generally, to induce high compatibility, a self-supervised task should:
(i) require the same features and
(ii) be as related to the \targettask{} as possible.
This ensures that the features that are important for the \targettask{} are made compatible and avoids conflicting objectives.
A counter example would be to use a reconstruction objective, which requires accurate localization, together with a classification objective, which is invariant to the exact location and instead requires semantic reasoning.

\section{Analysis of Compatibility: Additional experimental results}
\label{sec:additional_results}
In this section we provide additional analysis and experiments for Sec. 4 of our main paper.

\subsection{Using batch statistics at test time and other variants}
\label{sec:bn}
As noted in Sec. 4 of the main paper, 
our network components use Batch Normalization (BN)~\cite{ioffe15icml}.
By default, BN uses the statistics aggregated at training time for normalizing examples at test time.
This however leads to an incorrect normalization when recombining components.
In our main paper we therefore normalize features based on batch statistics of the test examples, in all our experiments. We explore variants here.

First of all, we tried the standard Batch Normalization which aggregates BN statistics over the training set. Next,
we tried recomputing BN statistics after recombination, by aggregating them over the complete test set, which primarily affects the BN statistics of the target task head.
Note that updating BN statistics requires only images, not labels.
Additionally, there exist alternative approaches to normalize feature statistics, which do not require any aggregated training statistics.
In particular, we experimented with Layer Norm~\cite{ba16arxiv} and Instance Norm~\cite{ulyanov16arxiv}. Finally, another way to make the feature representations of the feature extractors compatible, is to ensure they have a certain magnitude. We experimented with adding an L2 normalization layer after the feature extractors, which makes the features to become unit length. Additionally, we tried adding a loss to encourage these features to become unit length. Results are in Fig.~\ref{fig:batchnorm}.

\begin{figure*}
    \centering
    \includegraphics[width=\linewidth]{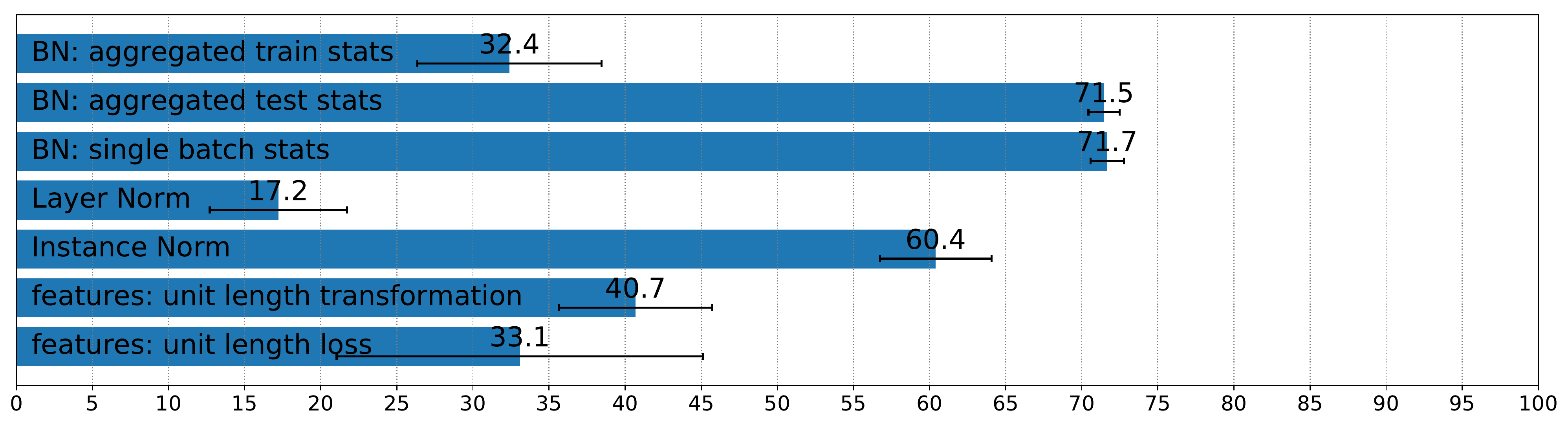}
    \caption{\textbf{Recombination accuracy for BatchNorm alternatives.} We explore different strategies for fixing the problem with unreliable BN statistics. We do this for \IIW+\DCC{}. Using batch-statistics at test time works best.}
    \label{fig:batchnorm}
\end{figure*}

First of all, we observe that using the normal aggregated training statistics works significantly worse than using batch statistics:
we only get 32.4\% recombination accuracy. This shows that it is important to use accurate statistics when recombining network components. 
Next, using the aggregated statistics over the whole test set and simply using statistics per batch perform best, outperforming alternatives by a significant margin.
Therefore we use single batch statistics at test time throughout our paper.

\subsection{Reaching the compatibility upper bound}
\label{sec:upper_bound}
To study our compatibility methods in a controlled setting, we repeat the analysis of the main paper (Sec. 4 %
), this time training both networks on the CIFAR-10 dataset. As the tasks are identical in this setting, the components of the two networks could, in principle, become \emph{perfectly compatible}.
In contrast, when the networks are trained on different two datasets, a classification head optimized for one is not expected to yield top accuracy on the other.

We show results in Fig.~\ref{fig:cifar10_cifar10}.
In this controlled setting, any combination of two or three of our compatibility methods allow to recombine network components into a new network $n_{ab}(\cdot)$ or $n_{ba}(\cdot)$ without a significant loss of accuracy: They all perform within 1.1\% of the upper bound of using the networks $n_{aa}(\cdot)$ and $n_{bb}(\cdot)$ directly. This shows that our methods are strong enough to achieve perfect compatibility, when the data allows it. In addition, advances in self-supervised learning could be used in our method to further improve this compatibility method
(See Sec.~\ref{sec:additional_discussion} above).

\begin{figure*}[th]
\centering
\includegraphics[width=1\linewidth]{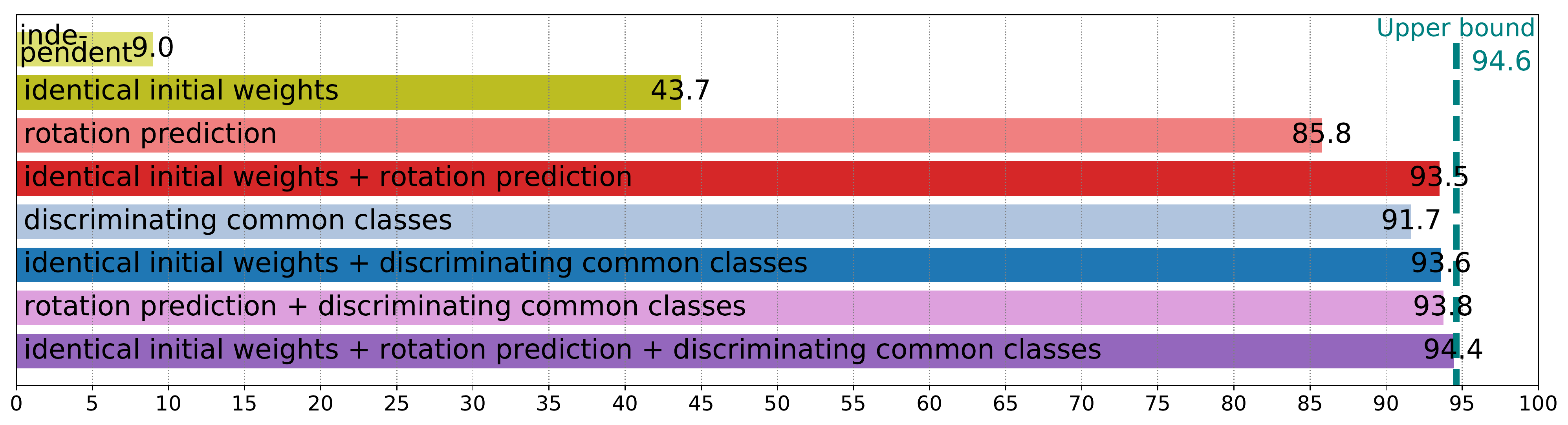}
\caption{
\textbf{Recombination accuracy for different methods when training both networks on CIFAR-10.} The numbers are an average over 3 runs.
}
\label{fig:cifar10_cifar10}
\end{figure*}

\subsection{Best layer for making features compatible}
\label{sec:layer_compatibility}
We study where to split the network into a feature extractor and a classification head in order to obtain the best compatibility for \IIW{}+\RP{} (Fig.~\ref{fig:layer_compatibility}). The results show that splitting at later layers leads to slightly lower \accuracy{} until the first block of stage 3. Splitting even later results in considerable drops in \accuracy.
This suggests that early features are the most compatible and re-usable across datasets, while mid-level features can also be made compatible quite well.
Instead, we hypothesize that late features are already highly specific to the trained network, and therefore harder to make compatible. Similar observations were made in analysis papers~\cite{li16iclr,mehrer18ccn,kornblith19icml}, where a recurrent result is that early network layers consistently learn the same features, while later layers learn increasingly specialized and different features, even if networks are trained on the same dataset.

\begin{figure*}[th]
\centering
\includegraphics[width=\linewidth]{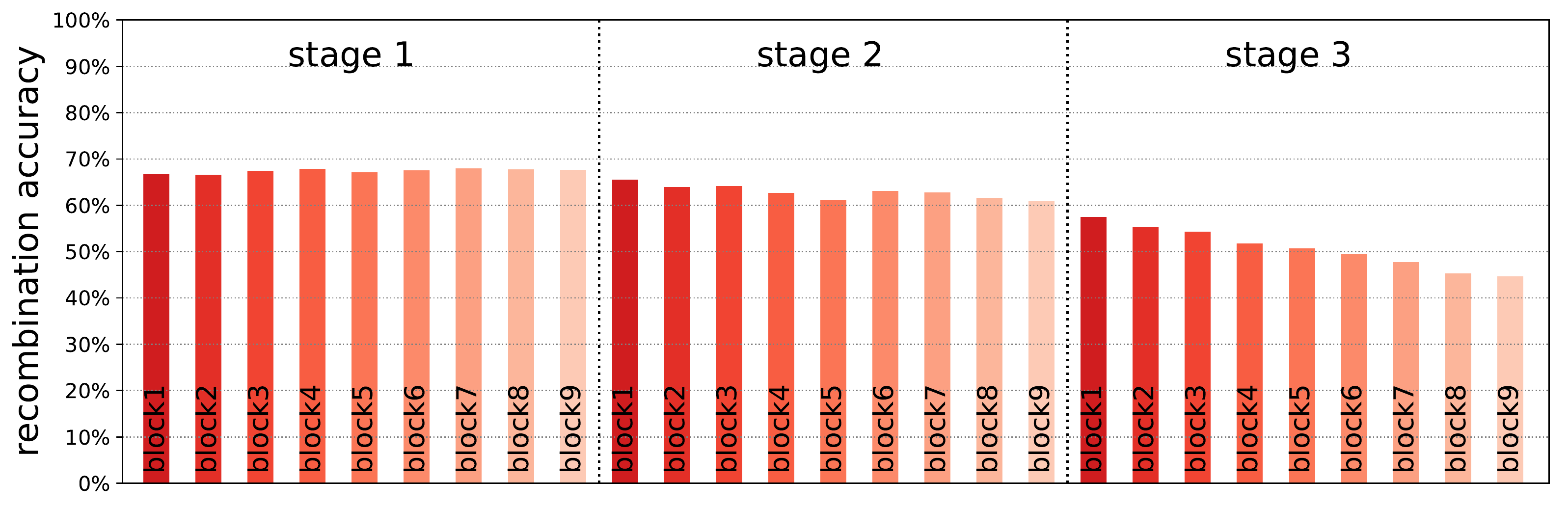}
\caption{
  \textbf{Effects of where to split the network.} We plot recombination accuracy averaged over 5 runs for \IIW{}+\RP{}, while varying the layer at which we split the network into a feature extractor and target task head.
}
\label{fig:layer_compatibility}
\end{figure*}

\subsection{Varying the number of common classes}
\label{sec:dcc_varying_classes}
When we use \DCC{} to produce compatible components, we use all 9 classes that STL-10 and CIFAR-10 have in common.
Here, we investigate what effect the number of common classes in the \DCC{} head has on performance, where we vary the classes used from 2 to 9.

Fig.~\ref{fig:varying_classes} shows results. We find that the number of classes in the \DCC{} head has a major effect on compatibility.
Going from 9 to 8 classes leads to a minimal effect on performance.
Instead, reducing to 6 classes or less has a strong negative effect on the \accuracy{} ($\leq 59.4\%$). In this regime, \DCC{} is outperformed by \RP{} (61.8\%).
Note that while \DCC{} trains only on the images with common classes, \RP{} always trains on all all images. It is unclear whether the drop in accuracy is caused by a lack of diversity in classes, or by using less training data.
We plan to investigate this in future work.

\begin{figure*}[t]
\centering
\includegraphics[width=1\linewidth]{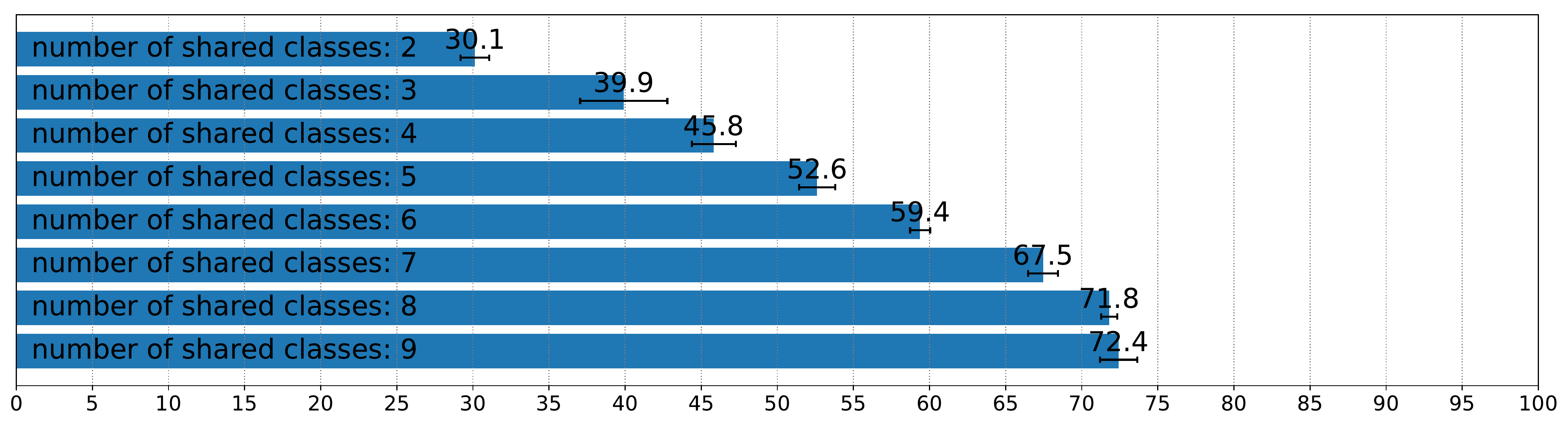}
\caption{
\textbf{Recombination accuracy when varying the number of common classes.} The numbers are an average over 3 runs for \IIW+\DCC{}, where we vary the number of classes on which the shared \DCC{} head is trained.
}
\label{fig:varying_classes}
\end{figure*}

\subsection{Per-dataset results for our analysis}
\label{sec:per_dataset_results}
In our main paper, we reported averages over the CIFAR-10 and STL-10 datasets (Fig.~2%
). For completeness and for better reproducability, we report also the per-dataset results in Fig.~\ref{fig:analysis_detailed}.
\begin{figure*}[t]
    \centering
    \begin{subfigure}{\linewidth}
    \includegraphics[width=\linewidth]{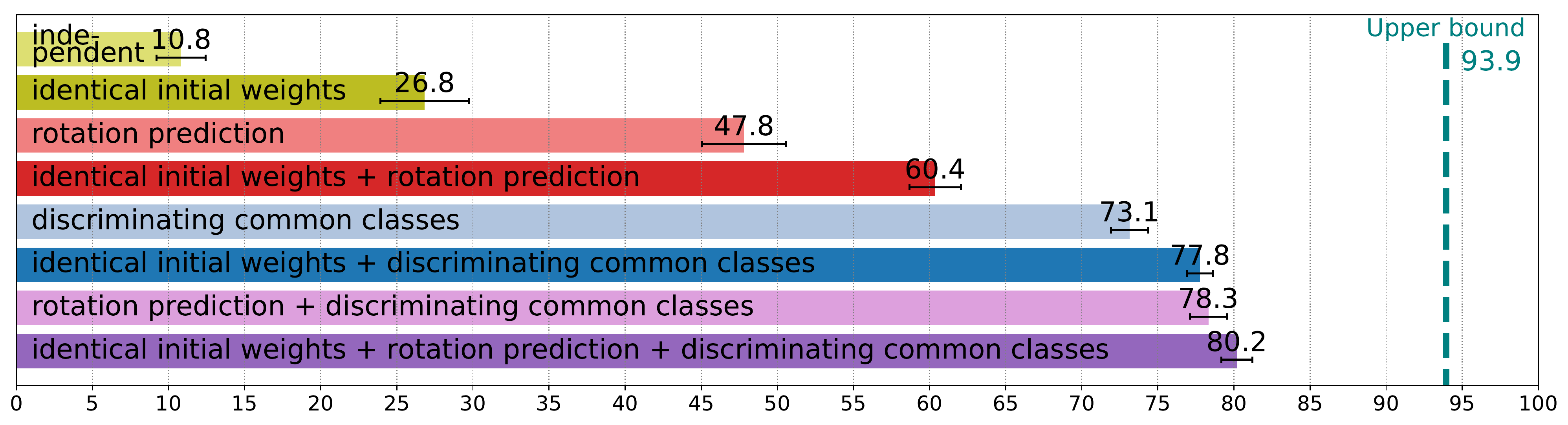}
    \caption{\textbf{Recombination accuracy, results on CIFAR-10 only.}}
    \end{subfigure}
    \begin{subfigure}{\linewidth}
    \includegraphics[width=\linewidth]{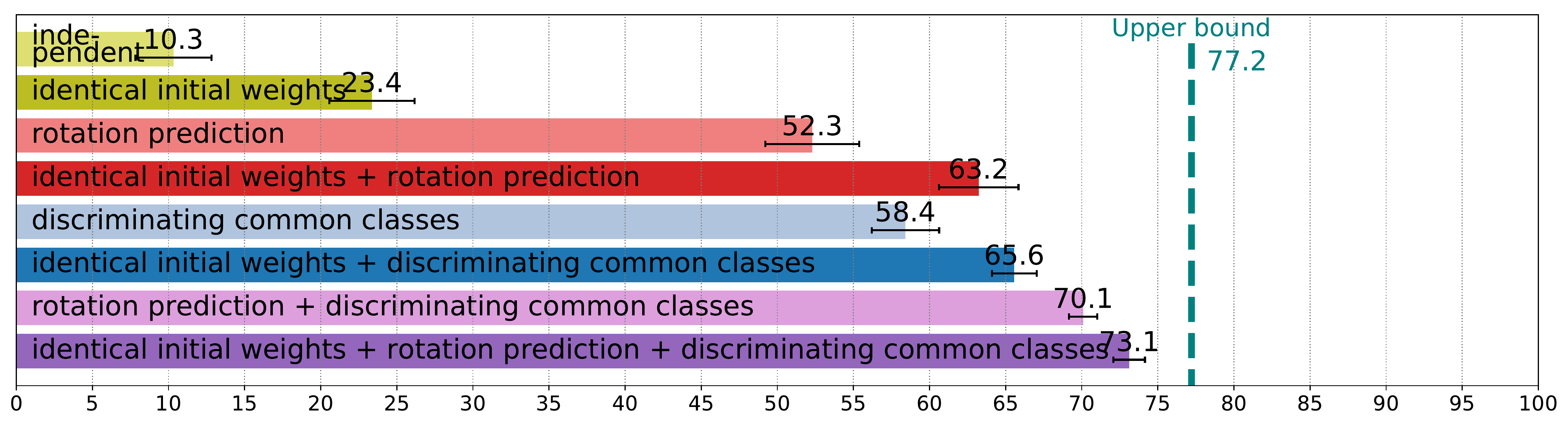}
    \caption{\textbf{Recombination accuracy, results on STL-10 only.}}
    \end{subfigure}
    \caption{\textbf{Recombination accuracy.} In this figure we show the individual results of CIFAR-10 and STL-10 (averages are in 		Fig.~2 of the main paper). %
		All numbers are averages over 10 runs, while the bars represent standard deviations.}
    \label{fig:analysis_detailed}
\end{figure*}

\subsection{Computational complexity \& Scalability}
Our methods require limited additional computation time, allowing to apply them to make many different networks compatible.
Specifically:
(i) IIW does not require additional computation time.
(ii) RP doubles the computational cost per task since each image needs to be processed twice (original and rotated version).
(iii) DCC is cheaper, as it only requires the features to be passed through an additional head.

In the number of tasks, our method scales linearly, as does training individual networks.
Hence, our method scales well computationally, allowing us to use it on several datasets (Sec. 5.3) or networks (Sec. 5.2 \& 5.3) at once.

\subsection{Classifier Transferability: Additional experimental results}
\label{sec:additional_classifier_transfer}
In Section 5.2 %
of our paper we demonstrated that we can transfer a classification head to multiple different backbones. In that experiment, we froze the feature extractor in the finetuning phase to maintain compatibility. Here, we also explore an alternative version. In particular, during the initial training phase (Fig. 1c), %
we add a (MobileNet V2) \RP{} head. During fine-tuning, we freeze this \RP{} head and use it encourage maintaining compatibility, while we allow the feature extractor to update its weights. Results of the orginal experiment (feature extractor frozen) and the just described variant (\RP{} frozen) are shown in Fig.~\ref{fig:multiple_backbones_variant}.

\begin{figure}[bt]
    \centering
    \includegraphics[width=1\linewidth]{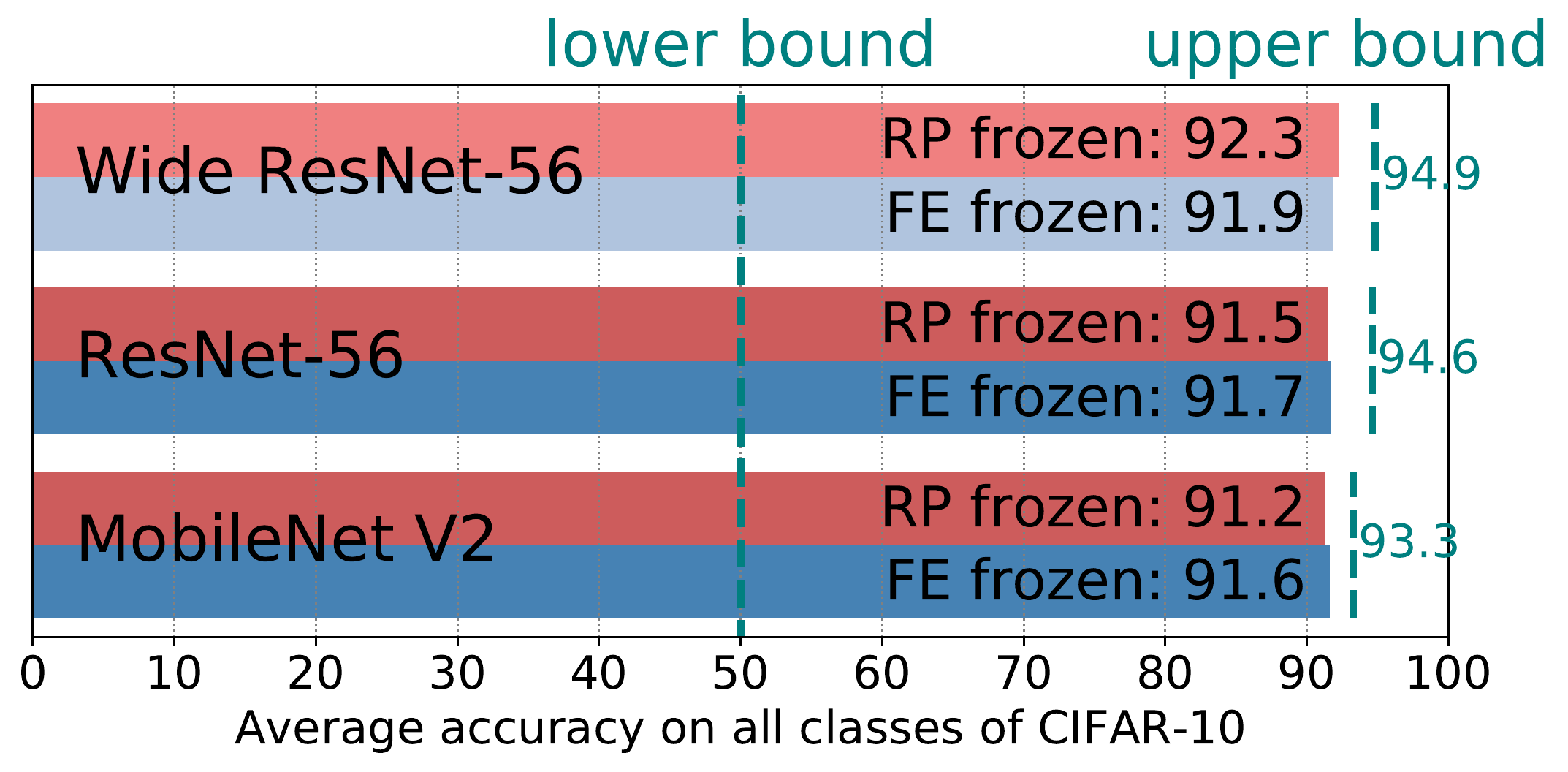}            
     \caption{\textbf{Accuracy when transferring classifier to compatible feature extractors.} }
    \label{fig:multiple_backbones_variant}
\end{figure}

We see that both variants achieve strong accuracy for all network component combinations. However, there is a trade-off:
freezing \RP{} allows the reference feature extractor to change, resulting in a higher accuracy for the reference combination. In contrast, freezing the reference \featureextractor{} ensures better compatibility with the other extractors, resulting in slightly higher accuracy for the non-reference recombinations.
Nevertheless, this shows that our method works well for maintaining compatibility, even when the parameters of the \featureextractor{} are allowed to change.

\section{Implementation details}
\label{sec:implementation_details}
\para{Resolution of the datasets.}
In order to have a consistent resolution we scale all datasets to $32\times32$, the resolution of the CIFAR-10 and CIFAR-100 datasets~\cite{krizhevsky09}.
We bilinearly downscale STL-10~\cite{coates11aistats}. 
Instead, for ILSVRC-12 classification~\cite{russakovsky15ijcv}, we use the $32\times32$ resolution version of~\cite{chrabaszcz17arxiv}.

\para{Hyperparameters.}
We investigated varying the relative weighting of the \targettask{} and the \auxiliarytask{} for \dcc{} and \rp{} and found that a large range of values works well. Hence, we set the weight for both of them to 1 for all experiments.
For all experiments we trained using an Adam optimizer~\cite{kingma15iclr} with a learning rate of $0.01$ and a weight decay of $10^{-7}$, unless mentioned otherwise.
We use a batch size of 512 images and train for 20'000 steps, where we decrease the learning rate by a factor of 10 every 5'000 steps.
These hyper parameters were set by reproducing the performance reported for a ResNet-56 trained and evaluated on CIFAR-10~\cite{he16cvpr}.
In the transfer learning experiments we use the same hyperparameters when training the initial models. When fine-tuning, we set the weight decay to zero for simplicity. We did not observe any significant performance difference, similar to~\cite{kolesnikov19arxiv}.

\para{Analysis of Compatibility: common classes of CIFAR10 and STL10.}
CIFAR-10 and STL-10 only have 9 out of 10 classes in common. Thus, while we always train on all 10 classes, we cannot evaluate on 10 when testing a CIFAR-10 classification head on the STL-10 test set (and vice-versa, see Fig.~1a in the main paper). Hence we ignore this class and only evaluate on 9 classes.

However, there is one exception: when we test whether the accuracy of the original task is compromised, we evaluate each network on the task which it is trained on and report accuracy on all 10 classes.

\para{Details on ``Starting from pre-trained models.'' (Sec. 4 of the main paper).}
For this experiment we start from models that were pre-trained for ILSVRC-12 classification~\cite{donahue13decaf,ren15nips,he2017mask}. To allow starting from either the same or \emph{different} pre-trained weights,
we produce two different pre-trained models by training from scratch twice, starting from different random initialization.
Furthermore, since STL-10 ~\cite{coates11aistats} contains ImageNet images, we create pre-trained models using self-supervised rotation prediction only (i.e. ignoring class labels).

\para{Unsupervised Domain Adaptation.} For the domain adaptation experiments we use a wide ResNet-28~\cite{zagoruyko16bmvc} of width 16,~\ie 3 stages of 4 blocks, where the stages consist of layers with 256, 512 and 1024 channels, respectively. This corresponds to the architecture used in~\cite{sun19arxiv_a}.
When training this model on the source, we train it with a batch size of 64 for 70'000 steps, reducing the learning rate by a factor of 10 every 20'000 steps.
As proposed in~\cite{zagoruyko16bmvc} we train our model using SGD with momentum~\cite{qian99nn} and a weight decay of $0.0005$.
We adapt the model to the target domain by fine-tuning for 1'000 steps with a learning rate of $0.00001$, the final learning rate of training on the source training set. During adaptation we set the weight decay at $0$.
As CIFAR-10 and STL-10 only overlap in 9 out of 10 classes, we train and test only on 9 classes, as is common in the literature~\cite{kumar18neurips,shu18iclr,lee19iccv,roy19cvpr,sun19arxiv_a}.

\fi

\end{document}